\documentclass[10pt,twocolumn]{article}

\usepackage[margin=0.75in]{geometry}
\usepackage{times}
\usepackage{microtype}
\usepackage{amsmath,amssymb,amsthm}
\usepackage{booktabs}
\usepackage{graphicx}
\usepackage{hyperref}
\usepackage{cuted}
\usepackage{caption}

\newtheorem{proposition}{Proposition}
\newtheorem{definition}{Definition}

\newcommand{\R}{\mathbb{R}}
\newcommand{\G}{\mathcal{G}}
\newcommand{\VG}{V_{\G}}
\newcommand{\VM}{V_M}
\graphicspath{{../figures/}}
\title{Manifold-GS: Certified Hybrid Assets via Varifold-Conservative Gaussian Splatting}
\author{Boyang Li\\Peking University\\\texttt{liboyang935@gmail.com}}
\date{Preliminary Draft}

\begin{document}
\maketitle

\begin{abstract}
3D Gaussian Splatting (3DGS) gives high-quality novel-view synthesis, but its
adaptive radiance primitives are not directly usable as structured assets:
opacity is not an additive area measure, clone/split operations can change the
induced geometry, and watertight mesh extraction can hallucinate collision
surfaces in unobserved regions.  We introduce \emph{Manifold-GS}, a certified
open-surface asset layer for Gaussian scenes.  The method separates appearance
opacity from geometric quadrature mass, interprets surface-like Gaussians as a
discrete unoriented varifold, and exports only confidence-certified open
surface patches while retaining uncertified content as a residual splat layer.
It provides refinement-conservative mass and moment transport rules, local
realizability diagnostics based on tangent and fundamental-form compatibility,
source-preserving patch bindings, and a hybrid bundle with certified geometry,
attached appearance Gaussians, residual Gaussians, and a conservative collision
candidate.  On three DTU scenes, a frozen asset-benchmark protocol shows zero
patch-defined certified edit leakage, patch texture round-trip PSNR of
30.1/35.3/33.7 dB, and
lower collision floater area than official 2DGS meshes on all scenes, with large
gaps on two scenes; it also reduces floater area relative to SuGaR culled meshes
and Poisson-from-3DGS under the same GT collision metric.  This precision comes
at lower coverage, so we present the result as a precision--coverage tradeoff
rather than a universal surface-reconstruction win.  External-region
annotations, phantom-collision probes, and 5k-face simplification further show
that the certified open-patch backbone preserves clean editing and collision
behavior under practical asset operations.  We also retain a negative finding:
local realizability is not sparse-RGB identifiability.  This draft anchors a
conservative claim: certified hybrid Gaussian assets, not state-of-the-art
RGB-only reconstruction.
\end{abstract}

\section{Introduction}

3D Gaussian Splatting (3DGS)~\cite{kerbl2023gaussian} optimizes anisotropic
Gaussian primitives for high-quality real-time novel view synthesis.  Its
adaptive density control makes the representation compact and expressive, but
also blurs the boundary between appearance fitting and geometric reconstruction.
A thin Gaussian may look like a local surfel, yet the collection of Gaussians
need not form a refinement-stable surface measure, a globally realizable normal
field, or an asset that can be edited, meshed, or bound to downstream geometry
tasks.

Recent surface-oriented Gaussian methods improve this situation by making
primitives planar~\cite{huang2024twodgs}, aligning Gaussians to mesh
surfaces~\cite{guedon2024sugar}, or introducing depth and normal
regularization~\cite{chen2024pgsr}.  Our starting point is complementary.  We
ask what geometric object is preserved by the adaptive Gaussian representation
itself.  In particular, if a Gaussian is split, cloned, pruned, or projected,
what should remain invariant so that geometry does not drift merely because the
renderer changed its sampling density?

We propose to treat surface-like Gaussians as samples of a discrete geometric
measure.  The covariance eigensystem provides a local tangent plane and
dimensionality label, while a separate positive mass variable plays the role of
area quadrature.  This mass is deliberately not identified with opacity, since
opacity participates in view-dependent alpha compositing and is not a conserved
geometric measure.  The resulting object is an unoriented varifold over
positions and tangent 2-planes.  This makes adaptive refinement a transport
problem: split and merge operations should preserve geometric mass and local
moments, up to controlled cell diameter terms.  It also gives an explicit source
mapping from original Gaussians to certified patch units, which is the basis for
structured editing and conservative collision proxies.

The second part of the paper separates two often conflated questions.  A local
tangent and curvature field may be \emph{realizable}, meaning that it is
compatible with some nearby surface, without being \emph{identifiable} as the
ground-truth surface from sparse RGB observations.  We therefore add
confidence-certified realizability diagnostics based on local support charts,
covariance normals, and fundamental-form compatibility, but we do not claim that
these diagnostics solve the inverse problem by themselves.

The contributions of this anchor draft are:
\begin{itemize}
  \item a refinement-conservative Gaussian-to-varifold interpretation that
  separates geometric mass from appearance opacity;
  \item conservation rules for adaptive Gaussian refinement and a stability
  statement in bounded-Lipschitz distance;
  \item a realizability diagnostic based on tangent support and approximate
  fundamental-form compatibility;
  \item a certified hybrid asset export that separates open surface patches,
  attached appearance Gaussians, residual splats, source mappings, and collision
  candidates;
  \item a frozen three-scene DTU asset benchmark covering certified editing,
  texture charting, collision precision, simplification robustness, external
  region annotations, and Poisson/SuGaR/2DGS mesh baselines;
  \item an experimental audit showing why sparse RGB-only compatibility should
  not be promoted to a surface-reconstruction claim.
\end{itemize}

\section{Related Work}

\paragraph{Gaussian scene representations.}
3DGS~\cite{kerbl2023gaussian} represents a scene by anisotropic 3D Gaussians
with learned opacity, covariance, and spherical-harmonic color.  Follow-up work
has improved aliasing, scale structure, efficiency, and level-of-detail
behavior~\cite{yu2024mipsplatting,wang2024octreegs}.  These methods retain the
strength of splatting as a real-time radiance representation, but the geometry
encoded by a trained Gaussian cloud is still not guaranteed to be a stable
surface.

\paragraph{Surface-oriented Gaussian splatting.}
2DGS~\cite{huang2024twodgs} replaces volumetric Gaussians by oriented disks,
making each primitive an intrinsic local surface element.  SuGaR~\cite{guedon2024sugar}
aligns Gaussians with surfaces and extracts/refines meshes.  Gaussian Opacity
Fields (GOF)~\cite{yu2024gof} and Sorted Opacity Fields~\cite{radl2025sof}
derive mesh extraction procedures from opacity fields rather than relying only
on external Poisson or TSDF fusion.  PGSR~\cite{chen2024pgsr},
VCR-GauS~\cite{chen2024vcrgaus}, and related depth/normal guided
methods~\cite{jia2025mndgs,lee2025prior} add geometric priors for more accurate
surface reconstruction.  MeshSplat~\cite{chang2026meshsplat} studies
generalizable sparse-view reconstruction through 2DGS-like surface primitives,
while GSurf~\cite{xu2026gsurf} couples Gaussian splatting with SDF learning for
continuous surfaces.  GeoSplatting~\cite{ye2025geosplatting} uses mesh guidance
to ground Gaussian normals for physically based inverse rendering.  Manifold-GS
does not claim priority over the broad idea of surface-aware Gaussian splatting.
Its narrower focus is the measure preserved by adaptive refinement, certified
open-surface asset export, and the boundary between local realizability and data
identifiability.

\paragraph{Geometric measures and compatibility.}
Varifolds provide an unoriented surface measure over positions and tangent
planes~\cite{simon1983gmt}.  Classical differential geometry relates the first
and second fundamental forms through compatibility conditions~\cite{doCarmo1976diffgeo}.
We use these tools as diagnostics for Gaussian scenes: if centers define a
support chart and covariances define normals and tangents, the two descriptions
should be mutually compatible before the result is trusted as a surface asset.

\section{Gaussian Splats as a Geometric Measure}

Let a trained Gaussian scene be
\begin{equation}
  \G = \{(\mu_i,\Sigma_i,\alpha_i,c_i)\}_{i=1}^N,
\end{equation}
where $\mu_i\in\R^3$ is the Gaussian center, $\Sigma_i$ is a positive-definite
covariance, $\alpha_i$ is opacity, and $c_i$ denotes appearance
parameters.  Let
\begin{equation}
  \Sigma_i
  = R_i \operatorname{diag}(\lambda_{i1},\lambda_{i2},\lambda_{i3})R_i^\top,
  \quad \lambda_{i1}\geq \lambda_{i2}\geq \lambda_{i3}.
\end{equation}
When $\lambda_{i1},\lambda_{i2}\gg\lambda_{i3}$, the two dominant eigenvectors
define a tangent plane $P_i$ and the smallest eigenvector defines an unoriented
normal $n_i$.  If $u_i,v_i$ are the two dominant eigenvectors, then
\begin{equation}
  P_i = u_i u_i^\top + v_i v_i^\top = I - n_in_i^\top .
\end{equation}

\begin{definition}[Gaussian varifold surrogate]
For surface-like Gaussian primitives with centers $\mu_i$, tangent planes $P_i$,
and positive geometric masses $q_i$, define
\begin{equation}
  \VG(\phi) = \sum_i q_i \phi(\mu_i,P_i),
\end{equation}
for bounded test functions $\phi$ over $\R^3$ and the Grassmannian of unoriented
2-planes.
\end{definition}

The key design choice is that $q_i$ is not $\alpha_i$.  Opacity is a rendering
parameter; it is neither additive under alpha compositing nor invariant under
view-dependent visibility.  In the current implementation, quantities such as
$\alpha_i\sqrt{\lambda_{i1}\lambda_{i2}}$ are treated as diagnostics or
initialization surrogates, not as a final definition of surface area.

The separation used throughout the paper can be summarized without a diagram:
\begin{equation}
(\mu_i,\Sigma_i,\alpha_i,c_i)
\quad\Longrightarrow\quad
\begin{cases}
(\mu_i,P_i,q_i) \mapsto q_i\delta_{(\mu_i,P_i)} & \text{geometry},\\
(\alpha_i,c_i) & \text{rendering}.
\end{cases}
\end{equation}
Here $P_i$ is extracted from the covariance eigensystem, while $q_i$ is a
separate geometric quadrature mass.  Opacity and color remain rendering
parameters.

\paragraph{Forward consistency.}

Let $M$ be a compact $C^2$ embedded surface with area measure $A$ and tangent
projector $P(x)$.  For a partition $\{C_i\}$, assume
\begin{equation}
  \sup_{x\in C_i}\|x-\mu_i\|\le h,\qquad
  \sup_{x\in C_i}\|P(x)-P_i\|_F\le \varepsilon_T,
\end{equation}
and $\sum_i |q_i-A(C_i)|\le \varepsilon_q$.  Then for any test function $\phi$
with supremum norm and Lipschitz constant bounded by one,
\begin{equation}
  |\VG(\phi)-\VM(\phi)|
  \le A(M)(h+\varepsilon_T)+\varepsilon_q .
\end{equation}
The proof is just cell-area quadrature plus the triangle inequality.  The point
is not that any trained 3DGS automatically satisfies these assumptions; it is
that once sampling, tangent error, and quadrature mass are controlled, the
covariance-to-varifold map has the expected forward consistency.
Proof details are given in Appendix~\ref{app:varifold}.

\section{Refinement Conservation}

Adaptive density control is central to Gaussian splatting.  From a geometric
measure perspective, however, a split should not create area merely because a
primitive was refined.

\begin{proposition}[Refinement conservation]
Consider replacing a parent primitive $i$ by children indexed by $j$ with
\begin{equation}
  \sum_j q_j = q_i,\qquad
  \sum_j q_j\mu_j = q_i\mu_i .
\end{equation}
Then zero-th mass and first spatial moment are preserved.  If additionally
$\sum_jq_jP_j=q_iP_i$, the tangent projector moment is preserved.
For any 1-Lipschitz $\phi$, the local perturbation satisfies
\begin{equation}
\begin{aligned}
\Delta_i(\phi)
&:=\biggl|\sum_jq_j\phi(\mu_j,P_j)-q_i\phi(\mu_i,P_i)\biggr|\\
&\le \sum_j q_j\bigl(\|\mu_j-\mu_i\|+\|P_j-P_i\|_F\bigr).
\end{aligned}
\end{equation}
\end{proposition}

This statement is intentionally conditional.  It does not say that arbitrary
3DGS optimization recovers a surface.  It says what an adaptive Gaussian method
should preserve if it wants its geometry to be meaningful across refinement.
The repository implements this idea through explicit geometric mass ledgers,
diagnostic projection, patch extraction, and conservative asset export.
The implemented conservative split recenters child offsets and inherits the
parent tangent, so mass, barycenter, and inherited-tangent moment are preserved
deterministically.  Merge preserves mass and barycenter, and preserves the
tangent moment only when child tangents agree.  Prune is handled as transport to
nearest retained atoms: if $\mathcal R$ is the removed set and $a(i)$ the
assigned retained atom, then
\begin{equation}
  \left\|\Delta\sum_i q_i\mu_i\right\|
  \le \sum_{i\in\mathcal R} q_i\|\mu_i-\mu_{a(i)}\|.
\end{equation}
This bound is recorded in the prune ledger.
Proof details and the prune transport derivation are given in
Appendix~\ref{app:varifold}.

\section{Realizability is not Identifiability}

The proposed compatibility diagnostics compare two local descriptions:
\begin{enumerate}
  \item support geometry fitted from Gaussian centers;
  \item extrinsic geometry predicted by covariance normals and their local
  derivatives.
\end{enumerate}
For a chart $\varphi(u,v)$ with Jacobian $J_\varphi$, the support induces a
first fundamental form $I=J_\varphi^\top J_\varphi$.  A support normal and a
covariance normal field induce two estimates of the second fundamental form.
Their disagreement, antisymmetry, Gauss residual, normal curl, and Codazzi-like
residual form a local realizability diagnostic.

This separation is the main reason the paper reports negative results.  A
smooth and compatible field can describe the wrong surface if the image evidence
does not constrain depth, support, or visibility sufficiently.  In the current
codebase, RGB-only training and fixed sparse support both expose this failure
mode.

Let $\Pi_ZV$ denote projection onto a compatible surface-varifold class $Z$.
For any varifold metric $d_V$,
\begin{equation}
  d_V(V,V_*) \le d_V(V,\Pi_ZV) + d_V(\Pi_ZV,V_*).
\end{equation}
The first term is a realizability error controlled by compatibility, sampling,
and quadrature quality.  The second term compares two already-realizable
surfaces and requires observation evidence; a smooth but wrong plane can make
the first term small while remaining far from the ground truth.  For the
implemented Gaussian kernel varifold, any coupling $\pi$ gives the sufficient
bound
\begin{equation}
  \operatorname{MMD}_k(\mu,\nu)
  \le
  \left[
  \int
  \left(
    \frac{\|x-y\|^2}{\sigma^2}
    +\frac{\|P-Q\|_F^2}{\tau^2}
  \right)d\pi
  \right]^{1/2}.
\end{equation}
Thus conservative transport controls the evaluator, but the reverse is not
guaranteed: a fixed-bandwidth MMD can hide small-scale geometric or topological
errors.  Reliable visible depth gives constructive coercivity because depth
$H^1$ error controls both position and tangent projector on visible regions.
RGB-only evidence is locally identifiable only when the restricted rendering
Jacobian has a strictly positive smallest singular value on the compatible
tangent space; low texture, occlusion, few views, and unobserved regions break
that condition.
The metric decomposition and kernel bound are expanded in
Appendix~\ref{app:identifiability}, while the observation assumptions are
summarized in Appendix~\ref{app:coercivity}.

\section{Implementation}

The current implementation is renderer-adjacent rather than a new rasterizer.
It includes:
\begin{itemize}
  \item PLY readers and covariance eigensystem diagnostics;
  \item Gaussian dimensionality labels and local patch graph scores;
  \item differentiable training hooks for thinness, support, tangent, shape,
  symmetry, and experimental Gauss compatibility terms;
  \item projection and patch-mesh extraction utilities;
  \item hybrid asset export with certified patches, attached Gaussians, residual
  Gaussians, source mappings, and a conservative collision candidate;
  \item observation-evidence gates based on sparse support and multi-view
  photometric consistency.
\end{itemize}

The implementation is designed to make unsupported geometry visible.  It keeps
uncertified regions as residual Gaussian layers rather than forcing a watertight
mesh over unknown areas.  The asset branch further includes DTU scan24/65/105
bundles, GLB packaging, collision and simplification metrics, restricted
rendering Fisher diagnostics, and external-region edit evaluation.

\paragraph{Frozen asset protocol.}
The downstream validation is intentionally not a screenshot-only mesh demo.  The
repository freezes a protocol, \texttt{asset-benchmark/1.0}, with explicit
PASS/FAIL checks for edit propagation, texture round-trip, and collision when
aligned GT is available.  The thresholds avoid quantities dominated by the
evaluation setup.  For editing, certified binding is expected to have zero leak
by construction, so the informative check is whether a proximity baseline
actually leaks across patch boundaries on the same scene.  For texture, the
absolute seam magnitude is dominated by true cross-patch color variation; the
gate therefore measures only the extra seam introduced by baking, together with
round-trip PSNR.  For collision, a single tight tolerance can report low
coverage when the tolerance is below the method's own geometric accuracy, so the
protocol reads coverage over a tolerance sweep and reports the tolerance at
which the coverage floor is reached.  Any threshold change is treated as a
protocol-version change.

\section{Experiments}

The following results are taken from the repository's current evidence ledger on
branch \texttt{a5-asset-benchmark-3scene}.  They should be read as an asset
utility and evidence-boundary audit, not as a final reconstruction leaderboard.
The experiments answer five questions: whether conservative operators prevent
refinement-induced measure drift; under what evidence compatibility is useful;
whether the certified subset yields a more reliable asset backbone; whether the
RGB-only failures match the identifiability analysis; and whether real-scene
increments replicate under a matched resource schedule.  We report Chamfer,
normal angle, kernel-varifold distance, certified mass/coverage, DTU
accuracy/completeness/overall, held-out PSNR/SSIM, edit leakage, texture
round-trip fidelity, collision floater area, phantom-collision rate, and
simplification robustness.  No single metric is used as a universal ranking.

\subsection{Mechanism: covariance and compatibility}

Unit tests verify that conservative clone/split preserves total mass,
barycenter, and inherited-tangent moment to numerical precision.  Pruning
preserves zero-th order mass through an explicit ledger and records the transport
cost and first-moment bound.  On the same geometry under different refinement
schedules, the explicit $q_i$ varifold diagnostic is more stable than treating
opacity as area.  These tests validate the operators, not reconstruction quality.

A 50-iteration synthetic smoke run verifies that the official 3DGS training path
can be instrumented and that the covariance-thinness loss is differentiable and
nonzero.  The median thinness diagnostic moves from approximately $0.371$ to
$0.299$, while $r_{23}$ moves from approximately $0.903$ to $0.716$.  This is
only evidence of a working thinness pressure.  It is not evidence of manifold
reconstruction.

On an analytic sphere checkpoint, offline projection substantially reduces
fundamental-form diagnostics: normal angle decreases from $21.84^\circ$ to
$0.96^\circ$, shape mismatch from $0.383$ to $0.113$, symmetry residual from
$0.364$ to $0.029$, Gauss residual from $0.300$ to $0.112$, normal curl from
$1.410$ to $0.024$, and Codazzi residual from $0.503$ to $0.057$.  Since the
projected field is constructed from the same MLS support, this is a diagnostic
sanity check rather than a reconstruction result.

On the sphere sparse-view asset mechanism test, the full compatibility pipeline
raises certified mass coverage from 54.95\% to 63.44\%, reduces certified point
Chamfer from 0.24090 to 0.15211, and reduces sampled patch-mesh Chamfer from
0.28287 to 0.16346, while all exported meshes have zero non-manifold edges.  The
cost is a small held-out appearance drop, from 21.4959 to 21.3542 dB PSNR and
from 0.5539 to 0.5395 SSIM.  The paired confidence interval does not clear the
predefined improvement threshold, so the formal verdict is \emph{inconclusive}.
We use it as mechanism evidence for asset extraction, not as a final
reconstruction claim.

\subsection{Analytic identifiability ladder}

Table~\ref{tab:ladder} shows the core decomposition.  Compatibility improves
tangent and varifold quality, but the oracle depth rung is needed for strong
support recovery.  The exact depth condition is an oracle diagnostic, not a fair
sparse-RGB method.

\begin{table}[t]
\vspace{-1.5mm}
\centering
\caption{Analytic plane, 1500-step mechanism ladder.  Lower is better except
PSNR.  RGB-only compatibility helps but does not close the gap to depth-anchored
geometry.}
\label{tab:ladder}
\begingroup
\scriptsize
\setlength{\tabcolsep}{3pt}
\begin{tabular*}{\columnwidth}{@{\extracolsep{\fill}}lrrrr@{}}
Method & Cham. & Norm. & Varif. & PSNR\\
\midrule
RGB & 0.05946 & 64.91 & 0.19335 & 28.017\\
RGB + comp. & 0.05011 & 55.18 & 0.15167 & 27.662\\
+ adapt. comp. & 0.04537 & 55.95 & 0.14614 & 30.264\\
+ depth oracle & \textbf{0.01885} & \textbf{8.84} & \textbf{0.06056} & \textbf{30.863}\\
\end{tabular*}
\endgroup
\vspace{-2mm}
\end{table}

\subsection{DTU matched diagnostic}

The current RGB-only manifold loss does not pass the frozen three-scene DTU
rule.  Against a matched 3DGS densification control on scan24, scan65, and
scan105, the mean relative overall improvement is only $0.271\%$, below the
$1\%$ threshold.  Accuracy improves on average by $1.98\%$, but completeness
degrades by $0.38\%$.  This supports the mechanism interpretation that the
current loss behaves more like a local precision regularizer than a reliable
surface-coverage improvement.

Adding a fixed RGB-SfM support anchor changes the picture.  Excluding scan105
as the post-hoc discovery scene, the frozen two-scene replication on scan24 and
scan65 improves the matched diagnostic overall metric by $1.62\%$ and $4.32\%$,
respectively, with held-out PSNR changes of $+0.132$ dB and $+0.297$ dB.  This
supports the narrower claim that sparse RGB-SfM support anchors and
realizability constraints can provide a reproducible increment under a matched
resource schedule.

\begin{table}[t]
\centering
\caption{DTU fixed RGB-SfM anchor replication.  Overall is the official DTU
diagnostic metric; lower is better.}
\label{tab:dtu-anchor}
\begingroup
\small
\setlength{\tabcolsep}{3pt}
\begin{tabular*}{\columnwidth}{@{\extracolsep{\fill}}lrrrr@{}}
Scan & 3DGS & Anch. & Gain & $\Delta$PSNR\\
\midrule
24 & 1.7163 & \textbf{1.6885} & +1.62\% & +0.132\\
65 & 2.4643 & \textbf{2.3577} & +4.32\% & +0.297\\
\end{tabular*}
\endgroup
\end{table}

\subsection{Mesh-vs-splat tradeoff with SuGaR}

A same-machine 8GB SuGaR pilot on DTU scan24/65/105 shows a tradeoff rather
than a single winner.  SuGaR gives stronger mesh geometry on scan24 and scan65,
especially completeness, while anchored 3DGS preserves held-out rendering
quality and performs better on the cleaner scan105 geometry.  Because this pilot
uses a reduced SuGaR budget, it is a diagnostic comparison and must not be used
to claim broad superiority over full-budget SuGaR.

\subsection{Three-scene certified asset benchmark}

The main asset result is a frozen CPU protocol, \texttt{asset-benchmark/1.0},
run on DTU scan24, scan65, and scan105.  Each exported bundle contains certified
open patches, a source-ID mapping, attached appearance Gaussians, residual
Gaussians, a conservative collision candidate, and a manifest.  The observation
gate is intentionally conservative: only about $42$--$58\%$ of patches, covering
roughly $50$--$62\%$ of surface area, are accepted; the rest remain as residual
splat content.

Coverage is treated as a recall-side diagnostic rather than a correctness score.
For collision assets, high coverage is useful only when paired with low false
surface area: watertight or TSDF extraction can cover more GT points by closing
unknown regions, while simultaneously introducing collision geometry in free or
unobserved space.  Thus a low-coverage result exposes the cost of certification,
whereas a high-coverage result is not sufficient evidence of a usable asset
unless its floater and phantom-collision rates remain low.

\begin{table}[t]
\centering
\caption{Certified asset summary on three DTU scenes.  Identification is
deliberately conservative; the remaining columns measure whether accepted
patches behave as usable assets.  Lower is better for floater, phantom, and
external non-target motion; higher is better for identified area and texture
PSNR.}
\label{tab:asset}
\begingroup
\scriptsize
\setlength{\tabcolsep}{3pt}
\begin{tabular*}{\columnwidth}{@{\extracolsep{\fill}}lrrrrr@{}}
Scan & ID area & floater & phantom & tex. dB & ext. move\\
\midrule
24 & 54.2\% & 18.25\% & 0.10\% & 30.11 & 0.36\%\\
65 & 49.5\% & 0.87\% & 0.00\% & 35.34 & 0.95\%\\
105 & 61.9\% & 1.54\% & 0.00\% & 33.70 & 0.66\%\\
\end{tabular*}
\endgroup
\end{table}

To avoid defining the edit target by our own patch IDs, an external operator
manually selected one contiguous region in each \texttt{certified\_patches.ply}
using Blender, based only on geometry appearance.  The selected regions contain
1,759/1,805/1,932 vertices for scan24/65/105.  Approximating each external
target by certified patches with at least $50\%$ overlap yields IoU
0.274/0.525/0.525 and precision/recall 0.619/0.329, 0.678/0.700, and
0.684/0.694.  This is not a claim of perfect semantic segmentation; scan24 in
particular exposes under-coverage at the patch granularity.  However, under a
rigid translation of $0.1\times$ the target bounding-box diagonal, certified
patch binding moves non-target source mass by only 0.36/0.95/0.66\%, with zero
residual contamination.  The nearest-radius baseline moves 22.3/43.9/19.2\%
non-target source mass and contaminates the residual layer by 24.2/45.4/20.4\%.
This is the non-circular edit evidence: the target is externally annotated, then
approximated by certified patches.  It supports conservative binding after
patch approximation, not zero-error semantic segmentation.

Texture gives a similar protocol lesson.  On scan105, noisy SH-DC colors give a
seam PSNR of 12.5 dB, while multi-view photometric mean colors raise it to
18.36 dB.  In both cases, the baked seam is close to the raw cross-boundary
color ceiling, so the remaining seam is mainly a color-source limitation rather
than a charting artifact.  A single-chart ablation is much worse even at matched
texel budget, trailing per-patch charting by 6--15 dB across scan24/65/105,
because one tangent chart cannot parameterize the curved object without
projection collapse.

\subsection{Collision precision against mesh baselines}

We compare the certified collision candidate to Poisson-from-3DGS, SuGaR native
culled meshes, and official 2DGS 30k TSDF meshes under the same DTU GT alignment
and collision-vs-GT evaluator.  The DTU STL mesh is deterministically mapped to
the Gaussian/reconstruction frame by the preprocessing \texttt{scale\_mat}; no
ICP is used.

\begin{table}[t]
\centering
\caption{Collision floater area at 1\% bbox tolerance.  Lower is better.
Coverage is lower for ours, reflecting a deliberate conservative
precision--coverage tradeoff.}
\label{tab:collision}
\begingroup
\scriptsize
\setlength{\tabcolsep}{3pt}
\begin{tabular*}{\columnwidth}{@{\extracolsep{\fill}}llrr@{}}
Scan & Method & Float. \% $\downarrow$ & Cov. \% $\uparrow$\\
\midrule
24 & ours & \textbf{18.3} & 37.1\\
24 & 2DGS & 21.1 & 69.9\\
24 & SuGaR & 78.7 & 69.8\\
24 & Poisson & 98.5 & 51.5\\
\midrule
65 & ours & \textbf{0.9} & 26.3\\
65 & 2DGS & 13.7 & 48.7\\
65 & SuGaR & 17.4 & 46.3\\
65 & Poisson & 97.4 & 49.0\\
\midrule
105 & ours & \textbf{1.5} & 41.0\\
105 & 2DGS & 11.4 & 51.8\\
105 & SuGaR & 7.7 & 51.9\\
105 & Poisson & 54.1 & 75.9\\
\end{tabular*}
\endgroup
\end{table}

This comparison is adequate for the collision-precision part of the asset claim:
2DGS is evaluated from its official 30k native TSDF meshes, under the same DTU
GT alignment, tolerance, and area-sampling evaluator as the other baselines.  The
result supports a narrow asset claim: certified open patches reduce false
collision surfaces compared with watertight or TSDF-style extraction, but they
cover less of the scene.  We therefore do not rank the method as universally
better than 2DGS, SuGaR, or Poisson; the methods occupy different points on a
coverage-versus-precision frontier.  Because the 2DGS mesh does not expose
certified patch/source binding, it is not used as a structured-editing baseline.
For the edit axis, the structural comparison is with watertight Poisson and
SuGaR meshes: their largest connected component occupies 97.5--99.5\% of
triangles, so sub-region editing has no natural certified boundary and falls
back to proximity selection, which is exactly the leaking baseline above.  Our
certified output instead exposes 381--402 observation-supported editable patch
units across the three DTU scenes.

\subsection{Simplification robustness and asset behavior}

To test whether the asset survives standard post-processing, we apply the same
Open3D cleanup and quadric decimation to ours, 2DGS, SuGaR, and Poisson,
targeting about 5k faces.  After simplification, our floater percentages remain
18.55/0.81/1.42 on scan24/65/105, compared with 18.87/12.21/10.12 for 2DGS,
78.89/17.16/7.46 for SuGaR, and 98.51/97.36/52.39 for Poisson.  Coverage drops
by only 1.0/0.2/0.8 percentage points relative to the original mesh, and all
outputs have zero non-manifold edges.  This supports robustness of conservative
collision precision after a common asset simplification step.  It does not
claim better triangle quality: 2DGS and Poisson have more regular triangles in
some mesh-quality statistics because our output intentionally preserves open
patch boundaries.

Two end-to-end demonstrations translate the metrics into behavior.  In a
phantom-collision probe, the fraction of free-space probes blocked by the
collision mesh is 0.00--0.10\% for ours, 0.05--3.58\% for SuGaR, and
2.90--6.50\% for Poisson.  In a scan105 semantic-part edit, rotating a selected
cluster of 30 certified patches moves no non-target source mass under certified
binding, whereas proximity binding drags 3,847 neighboring vertices (19.7\%).
The bundle is also packaged as a GLB with patch-colored certified geometry and a
transparent collision proxy node.

\begin{strip}
\vspace{-6pt}
\centering
\includegraphics[width=\textwidth,height=0.34\textheight,keepaspectratio]{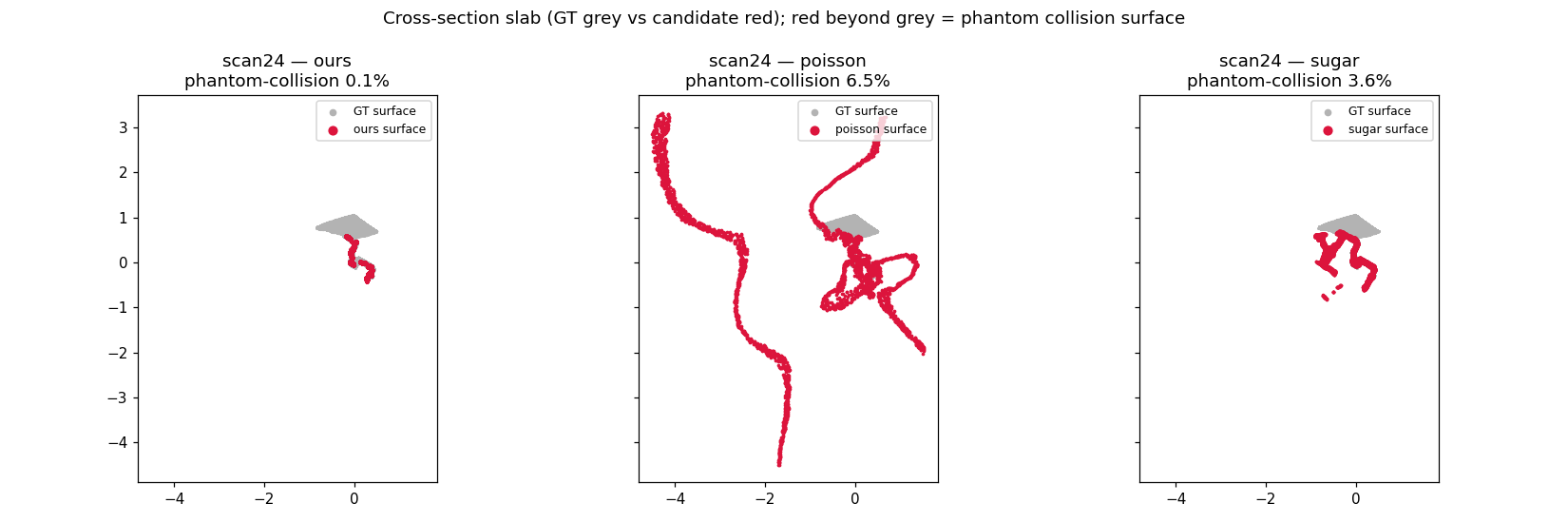}
\captionof{figure}{Collision cross-section demonstration on DTU scan24.  The row
visualizes the GT surface together with candidate collision geometry;
hallucinated geometry appears as free-space collision.  Across scan24/65/105,
the phantom-collision rate is 0.00--0.10\% for ours, 0.05--3.58\% for SuGaR,
and 2.90--6.50\% for Poisson.}
\label{fig:physics-demo}
\vspace{-8pt}
\end{strip}

\begin{strip}
\vspace{-6pt}
\centering
\includegraphics[width=\textwidth,height=0.34\textheight,keepaspectratio]{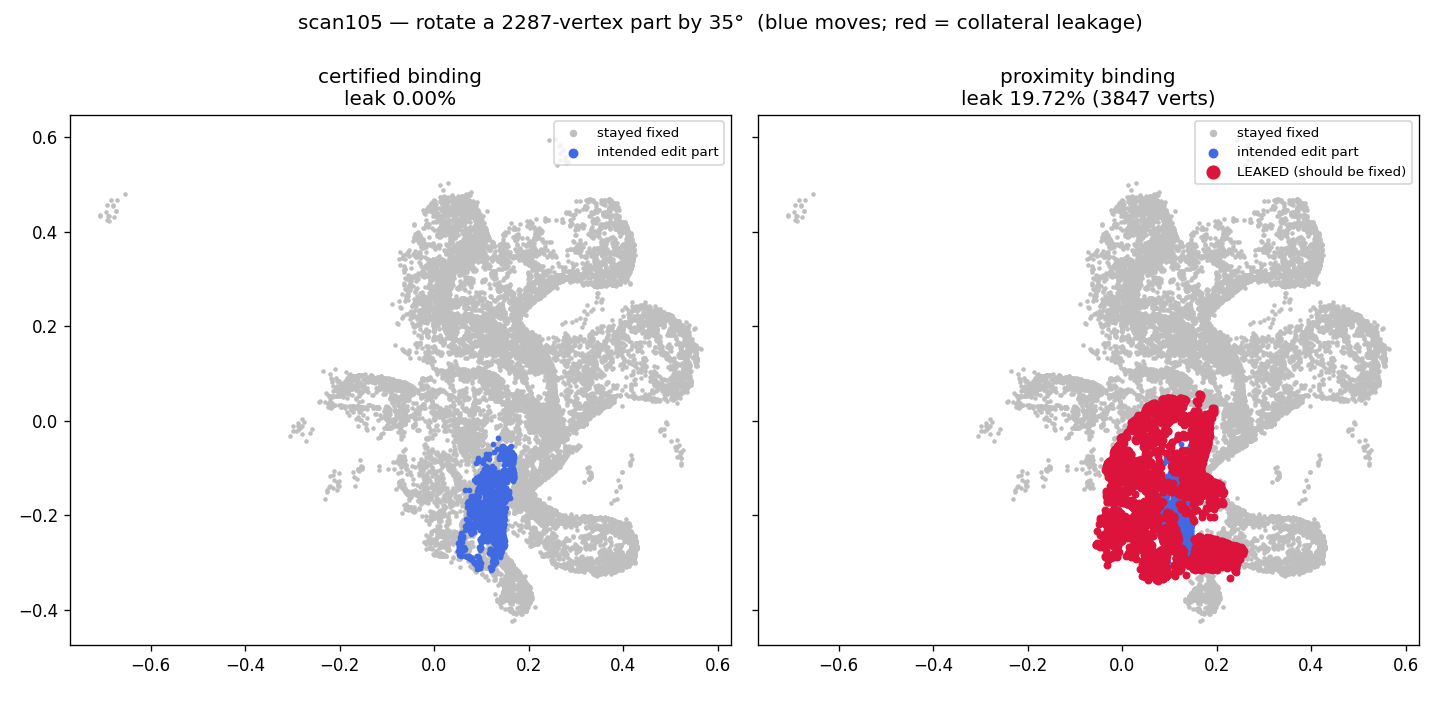}
\captionof{figure}{Certified edit propagation on scan105.  A selected part is transformed
through source-preserving certified patch binding, while a proximity selection
drags neighboring geometry.  Certified binding moves no non-target source mass;
the proximity baseline moves 3,847 neighboring vertices, or 19.7\% of the
source geometry.}
\label{fig:edit-demo}
\vspace{-8pt}
\end{strip}

\subsection{Restricted rendering Fisher diagnostics}

The asset branch also evaluates a restricted-rendering Fisher/Jacobian
diagnostic for accepted patches.  Under fixed appearance, 165/184 patches in
scan24, 146/164 in scan65, and 210/234 in scan105 are classified as locally
supported, with 19/17/24 weak patches and one insufficient-view patch in
scan65.  This certificate is a local RGB sensitivity ranking, not a proof of
global RGB-only reconstruction.  A GT read-only diagnostic confirms the boundary:
scan24 floater patches have lower Fisher median than clean patches, but scan105
floaters can still be marked supported.  The diagnostic therefore complements
collision-vs-GT evaluation; it does not replace it.

\section{Discussion}

The experiments separate three levels that are easy to conflate.  Thin
covariance is a primitive-level property; without a conserved $q_i$, the
collection-level geometry can still drift under densification.  Compatibility is
a field-level property; it rejects tangent and curvature fields that are not
locally surface-realizable, but it still permits many wrong yet realizable
surfaces.  Identifiability is an observation-level property; depth, RGB-SfM
support, or a locally coercive rendering Jacobian is needed to choose the GT
surface inside the compatible class.

This explains the DTU results.  Without an anchor, the loss improves local
accuracy but does not reliably improve completeness.  Fixed COLMAP support adds
a data-coercive signal in the coverage direction, making the same realizability
machinery beneficial on the two replication scenes.  The resulting 2.97\%
replication gain validates the decomposition; it is not the paper's main
performance claim.  The stronger claim is the asset-layer behavior: keep
certified geometry editable and collision-honest, and leave uncertain content as
residual splats rather than hallucinating a closed mesh.

\section{Limitations}

The current method has several important limitations.  First, sparse RGB-only
training does not reliably identify the ground-truth surface.  Second, local
patch charts and compatibility diagnostics do not guarantee global topology.
Third, the asset is intentionally conservative: it exposes open certified
patches and residual splats rather than producing a watertight mesh, so its
coverage is lower than 2DGS, SuGaR, or Poisson in several scenes.  Fourth, the
current evidence does not justify a state-of-the-art reconstruction claim
against full-budget surface-GS baselines.  Finally, the exported GLB is an asset
backbone, not a production-ready asset: complete UV atlas material baking,
renderer round-trip validation, engine-side physics validation, and richer
editable semantic units remain future work.

We also explicitly do not claim RGB-only superiority over 2DGS, SuGaR,
GeoSplatting, or other surface-GS systems; the unanchored DTU evidence argues
against that statement.  Nor do we claim unconditional convergence to the true
surface.  The theoretical statements are conditional on sampling, tangent,
mass, and observation evidence assumptions, while the engineering future work is
separate: full renderer round-trip PSNR/SSIM/LPIPS, complete UV-atlas material
baking, real-depth multi-seed validation, and full Gauss--Codazzi training.

\section{Conclusion}

Manifold-GS treats adaptive Gaussian splats as a conservative discrete geometric
measure and uses this view to build certified hybrid assets.  Separating
geometric mass from opacity makes refinement a transport problem rather than an
implicit change of surface area, while the varifold formulation provides a
common language for tangent-aware diagnostics and evaluator design.  The
exported asset keeps only observation-certified open patches as editable and
collision-relevant geometry, leaving uncertain content in residual splats rather
than forcing a watertight completion.

The experiments support this conservative asset interpretation.  Across three
DTU scenes, certified bindings avoid patch-defined edit leakage, per-patch
texture baking preserves round-trip appearance, collision candidates reduce
false surface area relative to mesh-extraction baselines, and the behavior
survives common simplification.  The same results also mark the boundary of the
method: certification lowers coverage, and local realizability does not by
itself identify the ground-truth surface from sparse RGB evidence.  Future
progress should therefore focus on stronger observation evidence and full
asset-rendering validation, rather than treating compatibility alone as a
complete reconstruction objective.

\newpage

\appendix

\section{Proof Details for the Varifold Bounds}
\label{app:varifold}

\subsection{Forward consistency}

Let $V_M$ denote the smooth surface varifold
\begin{equation}
  V_M(\phi)=\int_M \phi(x,P(x))\,dA(x).
\end{equation}
For a partition $\{C_i\}$ of $M$, insert the cell quadrature term
\begin{equation}
  \widetilde V(\phi)=\sum_i A(C_i)\phi(\mu_i,P_i).
\end{equation}
For any $\phi$ with $\|\phi\|_\infty\le 1$ and Lipschitz constant at most one,
\begin{align}
  |V_M(\phi)-\widetilde V(\phi)|
  &\le \sum_i \int_{C_i}
  |\phi(x,P(x))-\phi(\mu_i,P_i)|\,dA(x) \nonumber\\
  &\le \sum_i A(C_i)(h+\varepsilon_T)
   = A(M)(h+\varepsilon_T).
\end{align}
The mass error contributes
\begin{equation}
  |\widetilde V(\phi)-V_G(\phi)|
  \le \sum_i |A(C_i)-q_i|\,|\phi(\mu_i,P_i)|
  \le \varepsilon_q .
\end{equation}
Combining the two inequalities gives the bound in Sec.~3.  The result is a
forward consistency statement: it requires controlled sampling, tangent error,
and quadrature mass.  It is not a recovery theorem for arbitrary optimized
Gaussian scenes.

\subsection{Conservative refinement}

For one parent atom $(q,\mu,P)$ and children $(q_j,\mu_j,P_j)$, zero-th mass and
first spatial moment are exactly preserved by
\begin{equation}
  \sum_jq_j=q,\qquad \sum_jq_j\mu_j=q\mu.
\end{equation}
If $\sum_jq_jP_j=qP$, the tangent projector moment is also preserved.  For a
1-Lipschitz test function over position and projector space, define
$\Delta(\phi)=|\sum_jq_j\phi(\mu_j,P_j)-q\phi(\mu,P)|$.  Then
\begin{align}
  \Delta(\phi)
  &=\left|\sum_jq_j\bigl[\phi(\mu_j,P_j)-\phi(\mu,P)\bigr]\right| \nonumber\\
  &\le \sum_jq_j\bigl(\|\mu_j-\mu\|+\|P_j-P\|_F\bigr).
\end{align}
Summing over all refined parents gives the global bounded-Lipschitz perturbation
bound.  The implemented split recenters child offsets and inherits the parent
tangent, so the equalities hold deterministically for the inherited tangent
model.  Merge preserves tangent moment only in the special case of compatible
child tangents; otherwise it is a controlled approximation, not an exact
conservation operation.

Pruning is represented as transporting removed mass to retained atoms.  Let
$\mathcal R$ be removed atoms and $a(i)$ the retained assignment.  The zero-th
mass is preserved by adding $q_i$ to $a(i)$.  The first-moment change is
\begin{equation}
  \Delta\sum_i q_i\mu_i
  = \sum_{i\in\mathcal R}q_i(\mu_{a(i)}-\mu_i),
\end{equation}
hence
\begin{equation}
  \left\|\Delta\sum_i q_i\mu_i\right\|
  \le \sum_{i\in\mathcal R}q_i\|\mu_i-\mu_{a(i)}\|.
\end{equation}
This is the quantity recorded by the prune ledger.

\section{Realizability, Identifiability, and Kernel Distance}
\label{app:identifiability}

Let $Z$ be the class of compatible surface-varifold fields and let $\Pi_ZV$ be
any selected projection or approximation of $V$ in $Z$.  The metric decomposition
\begin{equation}
  d_V(V,V_*) \le d_V(V,\Pi_ZV)+d_V(\Pi_ZV,V_*)
\end{equation}
is only the triangle inequality, but it is conceptually important.  The first
term is the part that compatibility diagnostics can reduce.  The second term is
an observation problem: many compatible surfaces may explain the same sparse RGB
evidence.

For equal-mass positive measures, or after normalizing the compared measures to
equal total mass, let the Gaussian kernel varifold diagnostic use
\begin{equation}
  k((x,P),(y,Q))
  = \exp\!\left(-\frac{\|x-y\|^2}{2\sigma^2}
               -\frac{\|P-Q\|_F^2}{2\tau^2}\right).
\end{equation}
For any coupling $\pi$ between the normalized discrete varifold measures $\mu$
and $\nu$,
the reproducing-kernel distance is bounded by the transport cost
\begin{equation}
  \operatorname{MMD}_k(\mu,\nu)
  \le
  \left[
  \int
  \left(
    \frac{\|x-y\|^2}{\sigma^2}
    +\frac{\|P-Q\|_F^2}{\tau^2}
  \right)d\pi
  \right]^{1/2},
\end{equation}
up to the constant convention of the Gaussian kernel.  This follows from the
feature-map identity
\begin{equation}
  \|\Phi(a)-\Phi(b)\|_{\mathcal H}^2
  = k(a,a)+k(b,b)-2k(a,b)
\end{equation}
and $1-\exp(-t)\le t$.  Thus bounded spatial and tangent transport is sufficient
to control the implemented varifold evaluator.  The converse is not guaranteed:
a fixed-bandwidth kernel can smooth out small holes, thin handles, or
high-frequency tangent errors.

\section{Observation Coercivity Assumptions}
\label{app:coercivity}

Reliable visible depth gives a constructive identifiability condition.  On a
fixed visible chart $X(u,v)$, an $H^1$ depth error controls both position error
and first derivatives in the image-induced graph coordinates, and therefore
controls tangent projector error on visible regions under bounded camera
Jacobian and non-grazing-angle assumptions.  This is the reason depth or
calibrated sparse support can improve both position and tangent diagnostics.

RGB-only evidence has no such unconditional guarantee.  A local RGB solution is
identifiable only if the restricted rendering Jacobian, after quotienting out
appearance-only directions and restricting to compatible geometric perturbations,
has a strictly positive smallest singular value.  Textureless regions, repeated
patterns, occlusions, grazing views, and unobserved backsides can make this
singular value vanish or become too small to be useful.  This is why the paper
reports RGB-only failure rather than converting local compatibility into a
ground-truth reconstruction claim.

\bibliographystyle{plain}
\bibliography{references}

\end{document}